\documentclass[11pt]{article}
\usepackage[preprint]{acl}
\usepackage{times}
\usepackage{latexsym}
\usepackage[T1]{fontenc}
\usepackage[utf8]{inputenc}

\usepackage{microtype}
\usepackage{graphicx}
\usepackage{tikz}
\usepackage{subcaption}
\usepackage{booktabs}
\usepackage{amsmath}
\usepackage{amssymb}
\usepackage{mathtools}
\usepackage{amsthm}

\usepackage{multirow}
\usepackage{makecell}
\usepackage{array}
\usepackage{xcolor}
\usepackage[most]{tcolorbox}
\usepackage{xurl}

\ifdefined\shortpromptversion

\else

\fi

\title{
  Where Models Converge and Humans Diverge:\\
\makebox[\textwidth][c]{A Coverage Framework for Distributional Pluralism in Open-Ended Generation}
}
\author{
  Zini Yang\textsuperscript{1}\thanks{
    Currently a Ph.D. student at Johns Hopkins University,
    Baltimore, MD, USA.
    Email: \texttt{zyang181@jhu.edu}.
  }
  \quad
  Emily Wenger\textsuperscript{1,2}
  \quad
  Richard Jean So\textsuperscript{3}
  \\
  \textnormal{\textsuperscript{1}Department of Computer Science} \\
  \textnormal{\textsuperscript{2}Department of Electrical and Computer Engineering} \\
  \textnormal{\textsuperscript{3}Department of English} \\
  \textnormal{Duke University, Durham, NC, USA}
}
\begin{document}
\maketitle

\begin{abstract}

When a large language model (LLM) writes Harry Potter fanfiction, it can produce fundamental elements of the Hogwarts universe, such as recognizable places and characters. Human-written Harry Potter fanfiction, however, typically includes these fundamentals and much more, incorporating stylistically irregular content and relationship-diverse plotlines. This gap between LLM and human writing has been noted across a variety of domains. LLMs tend to produce “average” writing, while human writing contains more diverse content that covers a broader distribution. Existing work has shown the existence of this distributional “gap”. We take a further step by providing a systematic framework to measure coverage of the human response distribution and identify which regions model samples reach or miss.

Our paper proposes a human-grounded framework that uses the empirical distribution of human writing on a topic to measure the distributional breadth of LLM-generated content on that same topic. We propose two metrics, LLM Coverage (LLM-Cov) and In-Boundary Rate (IBR), which measure both distributional breadth and membership in the human reference region, respectively. Across ideation and narrative tasks, we find that the tested LLMs have lower mean coverage than equally sized held-out human samples. We also find that model reach is disproportionately concentrated near the center of the human response space relative to this human benchmark. In Harry Potter fanfiction, covered and uncovered passages also show systematic differences in canon anchoring and narrative form. Our framework can enable researchers to better assess the distributional breadth—what we term the "cultural reach"—of LLM-authored content.

\end{abstract}

\section{Introduction}

Generative AI has become a cultural technology. Large language models (LLMs) are trained on vast records of human expression and widely used to produce written content. When generating outputs, they make implicit choices about what counts as a recognizable genre, a plausible style, or a valid cultural form~\citep{brinkmann2023machine,farrell2025large}. This raises the question of how broadly LLM outputs reach across the range of responses produced by people and cultural communities.

This question is particularly relevant to open-ended generation tasks, for which there is no single correct answer. Two writers given the same prompt may draw on different backgrounds and interests to produce distinct stories that are both legitimate and appealing to readers. Harry Potter fanfiction provides a familiar example. Within this shared fictional world, writers explore different subgenres, relationships, and interpretations of canon, drawing on conventions within the fandom \citep{alfassi_fanfiction_2025}. Evaluating an LLM in this setting therefore involves asking how broadly its outputs reach across the range of stories written within this cultural community. We refer to this breadth as \emph{cultural reach}.

Pluralistic alignment offers a helpful way to think about this problem, asking whether LLMs can preserve heterogeneous human views, values, and preferences rather than collapsing them into an average response \citep{sorensen2024roadmap,xie2025survey,kirk2024prism}. Much existing empirical work measures pluralism through predefined stances, demographic groups, preference labels, or annotator disagreement \citep{santurkar2023opinions,meister2025distributional,novisdeutsch2025pluralist}. These approaches work well when the important dimensions of variation can be specified in advance. In open-ended generation, however, plurality also appears through different legitimate \emph{realizations} of the same prompt. Their differences may be semantic, stylistic, narrative, or community-specific, making the relevant dimensions difficult to specify exhaustively in advance.

Recent evaluations also derive opinions and perspectives from free-form human text \citep{mundada2026pluraleval,majer2026latent}, while distributional metrics systematically compare human and model writing \citep{le-bronnec-etal-2024-exploring,pillutla2021mauve}. This suggests an avenue for connecting statistical evaluation of LLM outputs to concrete patterns of LLM distributional coverage without requiring predefined categories. We take inspiration from such work in designing our approach. %, comparing human-generated content on a given subject to LLM-generated content on a similar topic. After examining the distributional elements LLMs fail to generate in a given domain, we can locate these elements in the human response space and examine the content and form of the missed texts. %This approach allows us to connect distributional evaluation to concrete patterns of LLM omissions without requiring predefined categories.

\paragraph{Our approach.}
We aim to understand the relative breadth of LLM and human-generated content, to gain a better sense of LLM's ``cultural reach'' on given open-ended topics. To do this, we evaluate the breadth of model generation against observed human responses. We introduce a geometry-based coverage framework that embeds human and model responses in a shared semantic space. We define the union of neighborhoods around human responses as an empirical human response boundary. We then use two metrics to compare the breadth of human- and LLM-generated content relative to these neighborhoods. \emph{LLM Coverage (LLM-Cov)} measures the fraction of reference human responses ``reached'' (e.g. distributionally close to) by LLM outputs. \emph{In-Boundary Rate (IBR)} measures how often LLM outputs fall inside the human response boundary. Finally, we use an equally sized, held-out human sample as a matched \emph{Human-to-Human (H2H)} reference to establish the coverage of a comparable human sample at the same sampling budget.

We use these metrics to compare human and LLM distributional reach on the Alternative Uses Task~\cite{guilford_nature_1967}, the Divergent Association Task~\cite{olson_naming_2021}, and Harry Potter fanfiction~\cite{alfassi_fanfiction_2025}. The tested LLMs exhibit lower mean coverage than matched human samples, consistent with prior findings of greater homogeneity in LLM responses ~\citep{wenger_were_2025}. We further find that this shortfall is largest at the periphery of the human distribution: model outputs reach central human responses more readily than peripheral ones. In a fanfiction case study, the human passages that models reach more often feature canonical characters, settings, and dialogue, while those they miss more often include characters from outside the Harry Potter universe and fragmentary prose.

\paragraph{Our contributions.}
\begin{enumerate}
    \item We develop a human-grounded, response-level coverage framework for open-ended LLM generations, using matched human samples to benchmark distributional breadth without predefined response categories.
    
    \item Across three tasks, the LLMs tested exhibit lower mean coverage than the human benchmark and tend to concentrate towards the center of the human distribution.% This finding provides a distributional interpretation of ``average'' generation.
    
    \item Our framework identifies the presence or absence of specific human response elements in LLM outputs, enabling textual diagnosis of coverage gaps. In HP fanfiction, for example, this analysis reveals key differences between LLMs and humans in character choice and narrative form. This diagnostic approach may enable future researchers to assess how changes in models, prompts, or decoding strategies enable generations that cover previously missed human responses.
\end{enumerate}

\section{Related Work}

\subsection{Pluralism and Cultural Context}
Pluralistic alignment asks how AI systems can reflect diverse human views, values, and preferences. \citet{sorensen2024roadmap} provide a framework distinguishing three forms of pluralistic alignment. \citet{xie2025survey} review training and inference methods for adapting models to different user preferences, while PRISM links participants’ backgrounds and preferences to their feedback on model conversations \citep{kirk2024prism}.

Assessing whether models reflect these differences requires comparing their generated responses with human data. Using public opinion surveys, \citet{santurkar2023opinions} find substantial discrepancies between model answer distributions and the opinions of different demographic groups. \citet{meister2025distributional} find that models are generally more accurate when directly reporting a population’s opinion distribution than when reproducing it through sampled responses. 

LLMs are now more commonly used for open-ended generation, including chatbot, raising the question of how to evaluate pluralism in their responses. \citet{brinkmann2023machine} examine machine generation as part of cultural production and transmission. In this setting, the forms of expression models repeatedly generate may influence which kinds of writing readers encounter. \citet{kommers2026computational} emphasize the importance of context and multiple reasonable interpretations when evaluating generated content. 

\subsection{Coverage of Opinions and Perspectives}

Several prior benchmarks have sought to measure whether LLM-generated content reflects the diversity of human opinions. For example, OvertonBench measures viewpoint representation through a set-coverage metric grounded in human judgments \citep{pooledayan2026overtonbench}. PluralEval evaluates open-ended responses against natural crowd discussions \citep{mundada2026pluraleval}. It extracts and clusters atomic opinions, then uses model judgments to assess which opinion clusters generated responses express. Finally, Majer et al.\ derive latent perspectives from book reviews by clustering sentence-level aspects and their review-level combinations \citep{majer2026latent}. They compare model outputs with equally sized held-out human samples using aspect and perspective coverage and distributional divergence.

In contrast, our evaluation targets how LLM outputs relate to individual human responses, including multiple realizations of similar content. This allows more precise analysis of narrative elements (e.g. events, relationships, expression) that LLMs may fail to generate, and does not require predefined categories for analysis. % Retaining these responses preserves their prevalence in the observed sample and keeps reached and unreached passages available for inspection. For narrative tasks, researchers can examine events, relationships, and expression alongside the nearest model response without first assigning each passage to a viewpoint category.

\subsection{Diversity, Homogenization, and Geometric Coverage}

An important direction in open-ended generation research is evaluating creativity. Creativity generally requires originality and effectiveness \citep{runco2012standard}, and its evaluation must consider the task and domain \citep{baer2005apt}. Recent work tests LLM creativity. \citet{chakrabarty_art_2024} evaluate the creativity of stories using expert-designed tests. CreativityPrism covers a broader range of tasks, including divergent thinking, creative writing, and reasoning, evaluating quality, novelty, and diversity separately \citep{hou_creativityprism_2025}. Meanwhile, \citet{jain_llm_2025} emphasize that evaluation should consider which kinds of diversity are meaningful for the task.

However, improvements in individual quality do not necessarily produce greater diversity across responses. \citet{doshi_generative_2024} find that AI-assisted stories receive higher creativity ratings but are also more similar to one another. \citet{anderson_homogenization_2024} find greater semantic similarity among ideas produced by participants using ChatGPT, while \citet{padmakumar_does_2024} find lower diversity in argumentative essays co-written with InstructGPT. Similar patterns appear in responses generated directly by models: using standardized creativity tasks, \citet{wenger_were_2025} find that LLM responses are more similar to one another than human responses are, \citet{zhang2026community} also find substantial variation in human preferences but relatively homogeneous model-generated responses.

Distributional metrics provide a way to compare generated samples against a human reference. MAUVE compares human and model text distributions in an embedding space, summarizing their differences in a single score \citep{pillutla2021mauve}. In image generation, \citet{kynkaanniemi2019improved} use $k$NN neighborhoods to estimate the support of real and generated samples and calculate precision and recall, and \citet{naeem2020reliable} build on this approach by introducing density and coverage metrics. \citet{le-bronnec-etal-2024-exploring} apply this geometric precision and recall idea to text generation to evaluate quality and diversity.

Our framework is closely related to these geometric approaches. We compare model and human response distributions, then examine where coverage gaps occur and what patterns the missed texts share. Our HP analysis provides a concrete example of how researchers can inspect which human expressions models miss. The framework can also help assess whether changes to models, prompts, or sampling strategies extend their coverage.

\section{Methodology}

\begin{figure*}[t]
    \centering
    \includegraphics[width=\linewidth]{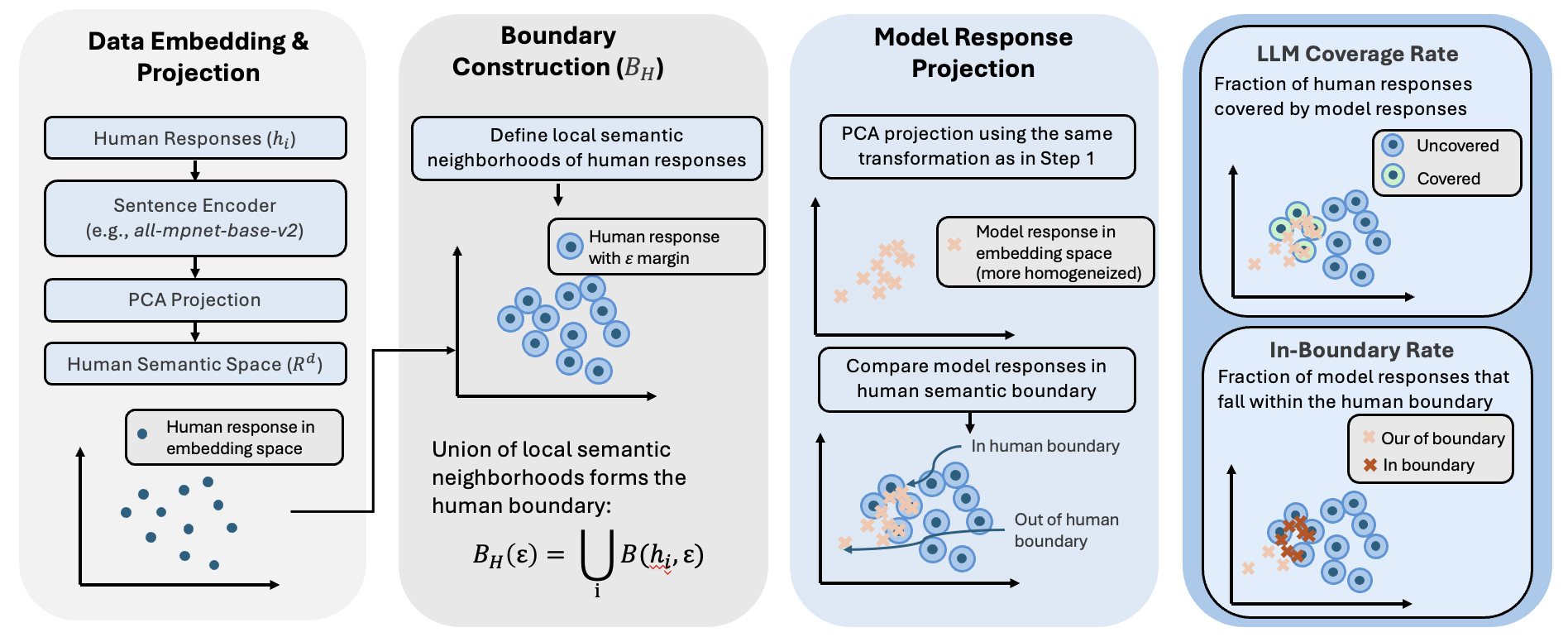}
    \caption{\textbf{Overview of our human-grounded coverage framework.}}
    \label{fig:flowchart}
    \vspace{-0.3cm}
\end{figure*}

Our framework evaluates the distributional breadth of open-ended generation by measuring which human responses model samples reach and which they miss. Figure~\ref{fig:flowchart} provides an overview of our approach.

\paragraph{Human Reference and Comparison Samples.}
For each evaluation setting, we use two non-overlapping human samples of equal size. The reference sample $H$ defines the human response region and supplies the responses to be covered. The held-out sample $H_{\mathrm{heldout}}$ provides the Human-to-Human comparison. We sample model responses $M$ with $|M|=|H_{\mathrm{heldout}}|$ and evaluate both sets against $H$ using the same evaluation pipeline.
% We repeat this procedure over 50 random splits of the human responses and report mean coverage across splits.

\paragraph{Data Embedding and Projection.}
We embed all reference human responses, held-out human responses, and model outputs using \texttt{all-mpnet-base-v2} \citep{reimers_sentence_2019,song_mpnet_2020}. We fit PCA only on the reference sample $H$ and retain the smallest number of components that explain at least 90\% of the variance. We then center and project $H_{\mathrm{heldout}}$ and $M$ using the same fitted transformation. Let
\[
H=\{h_i\}, \qquad H_{\mathrm{heldout}}=\{\tilde h_j\}, \qquad M=\{m_j\}
\]
denote the resulting projected embeddings.

\paragraph{Boundary Construction $(B_H)$.}
We approximate the empirical human response region as the union of local neighborhoods around the reference human responses. For each $h_i\in H$, let $r_i$ be its Euclidean distance to the $k$-th nearest point in $H\setminus\{h_i\}$. We define a global neighborhood radius as
\[
\varepsilon=Q_q(\{r_i\}),
\]
where $Q_q$ denotes the $q$-th quantile of the reference-point distances. The empirical human response boundary is then
\[
B_H(\varepsilon)=\bigcup_i B(h_i,\varepsilon),
\]
where $B(h_i,\varepsilon)$ is the closed Euclidean ball centered at $h_i$ with radius $\varepsilon$.

A response $x$ lies inside this boundary if it falls within $\varepsilon$ of at least one reference human response:
\[
x\in B_H(\varepsilon) \quad\Longleftrightarrow\quad \min_{h_i\in H}\|x-h_i\|_2\leq\varepsilon.
\]

We use $k=15$ and $q=0.5$, setting the radius to the median reference-human $k$-NN distance. We evaluate alternative quantiles and local radii to assess the sensitivity of this choice, and use a pairwise audit to validate that the induced matches correspond to meaningful task-relevant textual similarity.

\paragraph{Human-Grounded Coverage Metrics.}
We evaluate model outputs via two complementary metrics. Let $d(x,S)=\min_{s\in S}\|x-s\|_2$ denote the distance from $x$ to its nearest neighbor in $S$. Our first metric, {\bf LLM Coverage (LLM-Cov)}, measures the fraction of reference human responses reached by at least one model output:
\[
\mathrm{LLM\text{-}Cov}=\frac{|\{h_i\in H:d(h_i,M)\leq\varepsilon\}|}{|H|}.
\]
A reference human response is considered covered when at least one model output falls within its $\varepsilon$-neighborhood.

Our second metric, {\bf In-Boundary Rate (IBR)}, measures the fraction of model outputs that fall inside the empirical human boundary:
\[
\mathrm{IBR}=\frac{|\{m_j\in M:m_j\in B_H(\varepsilon)\}|}{|M|}.
\]

LLM-Cov asks how much of the reference human response space the model reaches. IBR asks how often model outputs fall within the empirical range of human responses. Figure~\ref{fig:coverage-geometry} provides an example from our task.
\begin{figure}[t]
\centering
\small
\setlength{\tabcolsep}{3pt}
\renewcommand{\arraystretch}{1.12}
\begin{tabular}{@{}lp{0.49\linewidth}l@{}}
\toprule
\textbf{Unit} & \textbf{Rope-use response} & \textbf{Matched to} \\
\midrule
\multicolumn{3}{@{}l}{\textbf{Human references}} \\
$h_1$ & ``Lasso cattle.'' & $m_1$ \\
$h_2$ & ``Lasso cows and steer.'' & $m_1$ \\
$h_3$ & ``Create a swing.'' & $m_2$ \\
$h_4$ & ``Wear it as a belt.'' & --- \\
\midrule
\multicolumn{3}{@{}l}{\textbf{Model outputs}} \\
$m_1$ & ``Lasso cattle.'' & $h_1,h_2$ \\
$m_2$ & ``Create a swing.'' & $h_3$ \\
$m_3$ & ``Use it as a lash.'' & --- \\
\bottomrule
\end{tabular}
\smallskip

\caption{\textbf{Illustration of how we measure coverage.}
For the responses presented in the table, $\mathrm{LLM\text{-}Cov}=3/4$, because $3/4$ human responses map to a model response, and $\mathrm{IBR}=2/3$, because $2/3$ model responses fall within the human response boundary.}
\label{fig:coverage-geometry}
\end{figure}
\vspace{-0.24cm}

\paragraph{Human-to-Human Reference.}
We compute H2H-Cov and H2H-IBR by replacing $M$ with $H_{\mathrm{heldout}}$ in the definitions above. We also report relative coverage, which expresses model coverage as a fraction of the matched human benchmark:
\[
R=\frac{\mathrm{LLM\text{-}Cov}}{\mathrm{H2H\text{-}Cov}}.
\]
We use this ratio to assess a model’s ability to cover the human reference relative to human sampling at the same budget.

\section{Evaluation Procedure}
\label{sec:evaluation-procedure}

Here, we describe the tasks, models, and metric implementations used for our experiments.

\vspace{-0.1cm}
\subsection{Tasks}
\label{sec:datasets}
\vspace{-0.1cm}

We evaluate coverage across three open-ended generation settings: the Alternative Uses Task (AUT), the Divergent Association Task (DAT), and Harry Potter fanfiction writing.
\textbf{AUT} is a widely used divergent-thinking task that asks participants to generate alternative uses for common objects, capturing ideational flexibility and originality \citep{guilford_nature_1967}. AUT provides a baseline for coverage in constrained ideation.
\textbf{DAT} elicits semantically distant concepts and is grounded in associative theories of creativity \citep{mednick_associative_1962,olson_naming_2021}. DAT provides a lexical association setting for evaluating coverage.
\textbf{HP Fanfic}, in contrast, provides a community-structured narrative setting in which legitimate responses can vary in characters, tone, relationship to canon, and fandom conventions. Recent work on fanfiction and AI highlights the relevance of fan communities, authenticity, and creative norms in this domain~\citep{alfassi_fanfiction_2025}. 

% Together, these tasks span constrained ideation, lexical association, and community-structured narrative writing. This allows us to examine coverage across different task settings, and fanfiction specifically provides a context for interpreting the textual character of the gaps.

\vspace{-0.1cm}
\subsection{Models Evaluated}
\label{sec:models-setup}

We evaluate five open-source instruction-tuned models: Qwen2.5-7B, Ministral-8B, Llama-3.1-8B, Qwen2.5-32B, and Mistral-Small-24B. Each model is evaluated at temperatures $t\in\{0.3,1.0\}$, giving ten configurations, with other decoding parameters held fixed unless otherwise stated.

% We also examine whether combining model outputs expands coverage. On a fixed human reference sample, we add one model's outputs at a time, choosing the model that adds the most coverage. Separately, we compare a mixture of all ten configurations with individual configurations, keeping the total number of outputs the same.

\vspace{-0.1cm}
\subsection{Metric Implementations}

We embed all human and model responses using \texttt{all-mpnet-base-v2} from SentenceTransformers\citep{song_mpnet_2020}.
We fit PCA only on the human reference sample and retain the minimum number of components that explain 90\% of its variance.
We define local neighborhoods using $k$-NN with $k=15$ and set the global radius $\varepsilon$ as the $50^{\text{th}}$ percentile of reference-human $k$-NN distances. Coverage estimates can vary with the human reference sample. We therefore average results over 50 random reference/held-out splits, refitting PCA and radii on each reference set.
%  Appendix~\ref{app:sample-ledger} gives the task-specific sample counts, assignment seeds, and analysis units.

For statistical testing, we use two-sided paired tests when comparing coverage metrics across matched model configurations.
For the descriptive HP covered-versus-uncovered analysis, we use Mann--Whitney U tests for continuous variables and two-sided chi-square tests for binary variables.
We report observed group differences and raw $p$-values in the paper.
% ; percentile bootstrap intervals for the principal HP contrasts are reported in Appendix Table~\ref{tab:hp-bootstrap-intervals}.

\section{Experimental Results}
\label{sec:experiments}

Next, we present the main results from our coverage-based evaluation of distributional pluralism on the three open-ended generation tasks.

\subsection{Overall Coverage}
\label{sec:coverage-results}

\begin{table*}[t]
\centering
\small
\setlength{\tabcolsep}{5.8pt}
\renewcommand{\arraystretch}{1.08}
\begin{tabular*}{\textwidth}{@{\extracolsep{\fill}}llcccccc@{}}
\toprule
\textbf{Temp.} & \textbf{Model}
& \multicolumn{2}{c}{\textbf{AUT}}
& \multicolumn{2}{c}{\textbf{DAT}}
& \multicolumn{2}{c}{\textbf{HP Fanfic}} \\
\cmidrule(lr){3-4}\cmidrule(lr){5-6}\cmidrule(lr){7-8}
& & \textbf{IBR} & \textbf{LLM-Cov}
& \textbf{IBR} & \textbf{LLM-Cov}
& \textbf{IBR} & \textbf{LLM-Cov} \\
\midrule
$t{=}0.3$ & Qwen2.5-7B & .996 & .095 & 1.000 & .465 & .995 & .459 \\
& Ministral-8B         & .993 & .242 & .926 & .117 & .970 & .322 \\
& Llama-3.1-8B         & .942 & .214 & 1.000 & .223 & .984 & .380 \\
& Qwen2.5-32B          & .989 & .207 & 1.000 & .205 & .993 & .428 \\
& Mistral-24B          & .934 & .282 & 1.000 & .090 & .925 & .406 \\
& Average              & .971 & .208 & .985 & .220 & .973 & .399 \\
\midrule
$t{=}1.0$ & Qwen2.5-7B & .947 & .372 & 1.000 & .666 & .976 & .553 \\
& Ministral-8B         & .901 & .622 & .860 & .448 & .943 & .519 \\
& Llama-3.1-8B         & .897 & .519 & .994 & .635 & .936 & .545 \\
& Qwen2.5-32B          & .953 & .349 & .999 & .556 & .983 & .536 \\
& Mistral-24B          & .886 & .614 & .993 & .456 & .955 & .520 \\
& Average              & .917 & .495 & .969 & .552 & .959 & .534 \\
\midrule
& \textbf{Overall average} & .944 & .352 & .977 & .386 & .966 & .467 \\
& \textbf{Human-to-Human}  & .900 & .889 & .930 & .894 & .899 & .885 \\
\bottomrule
\end{tabular*}
\caption{\textbf{Model coverage versus the human reference.}
Entries are means over 50 repeated random splits with equal model and held-out human sample sizes under the global $p50$ radius.}
\label{tab:coverage-results}
\end{table*}

The evaluated LLMs generally produce in-boundary responses but reach a narrower portion of the human reference distribution. Every tested configuration has lower mean LLM-Cov than the matched Human-to-Human reference across all three tasks (Table~\ref{tab:coverage-results}).

We find no consistent coverage advantage for larger models within the evaluated families. Increasing temperature from $t{=}0.3$ to $t{=}1.0$ raises mean LLM-Cov on every task, while mean IBR declines, most markedly on AUT. Higher-temperature sampling thus broadens model reach, but some generations then fall outside the human reference region, making them arguably less useful.

Mean relative coverage ($R$) across 50 splits remains below 1 at every tested global-radius percentile in each task (Table~\ref{tab:radius-r-main}), indicating that model coverage falls short of the human--human baseline on average across the evaluated neighborhood scales.

\begin{table}[t]
\centering
\small
\setlength{\tabcolsep}{7pt}
\renewcommand{\arraystretch}{1.0}
\begin{tabular}{lccc}
\toprule
\textbf{Radius percentile} & \textbf{AUT} & \textbf{DAT} & \textbf{HP} \\
\midrule
$p25$ & .205 & .273 & .355 \\
$p50$ & .396 & .432 & .527 \\
$p75$ & .595 & .627 & .700 \\
\bottomrule
\end{tabular}
\caption{\textbf{Relative coverage across neighborhood scales.}
We report $R=\mathrm{LLM\text{-}Cov}/\mathrm{H2H\text{-}Cov}$}
\label{tab:radius-r-main}
\end{table}

\begin{figure}[t]
\centering
\includegraphics[width=\linewidth]{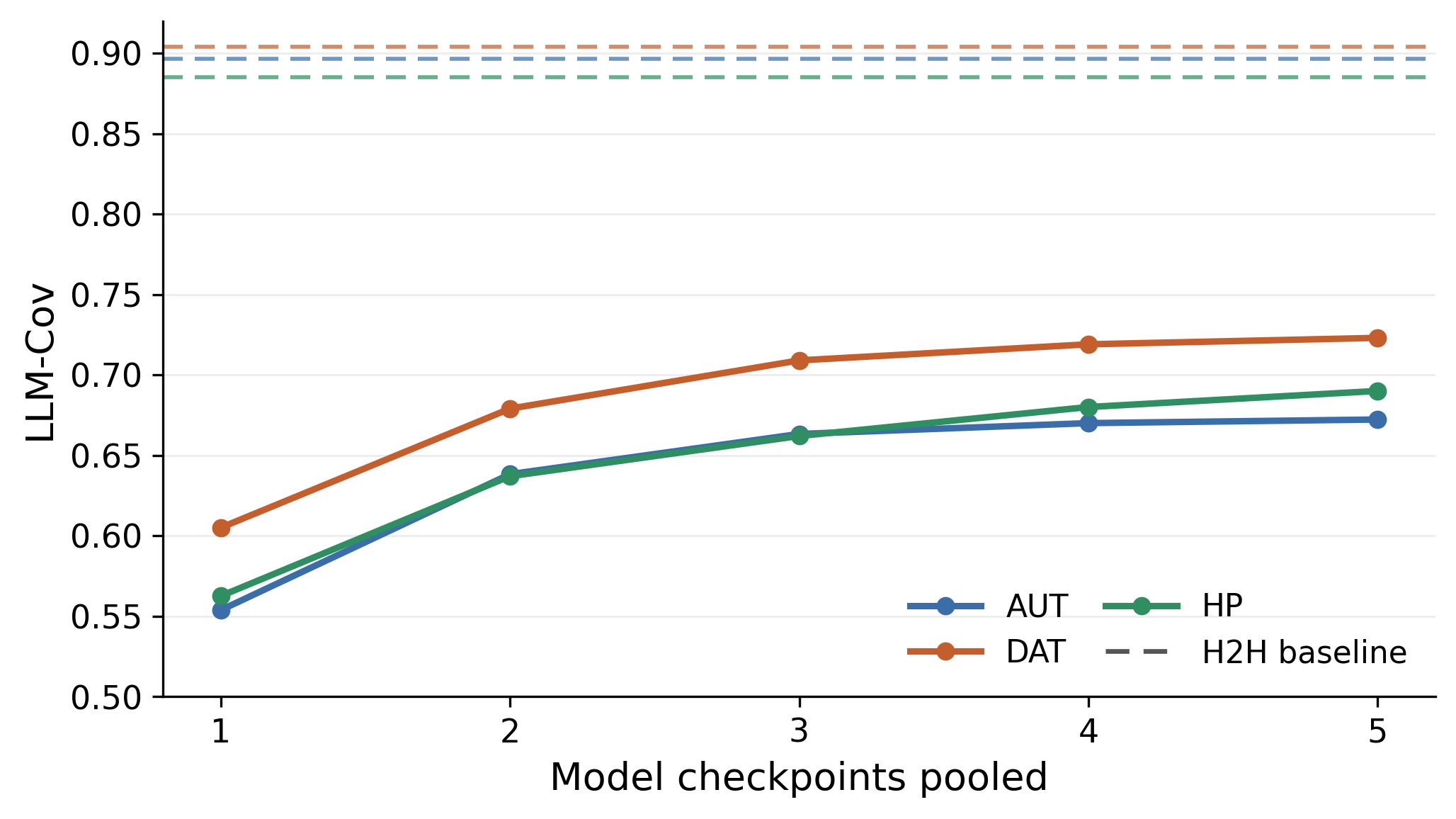}
\caption{\textbf{Sequential model addition yields diminishing coverage gains.}}
\label{fig:sequential-addition}
\end{figure}

\paragraph{Combining models.}
Figure~\ref{fig:sequential-addition} also shows diminishing coverage gains as outputs from additional models are pooled. Despite the growing number of model responses, later models reach relatively few previously uncovered human responses, and the combined pool remains below the fixed-size H2H reference.

\subsection{Model Coverage Favors Central Responses}
\label{sec:centrality-coverage}

We next examine which parts of the human response distribution model samples reach. Some human responses lie close to many other examples, while others occupy more peripheral regions. We rank reference responses into ten deciles using a geometric centrality score, from central to peripheral (Appendix~\ref{app:centrality-definition}).

\begin{figure*}[t]
\centering
\includegraphics[width=.90\textwidth]{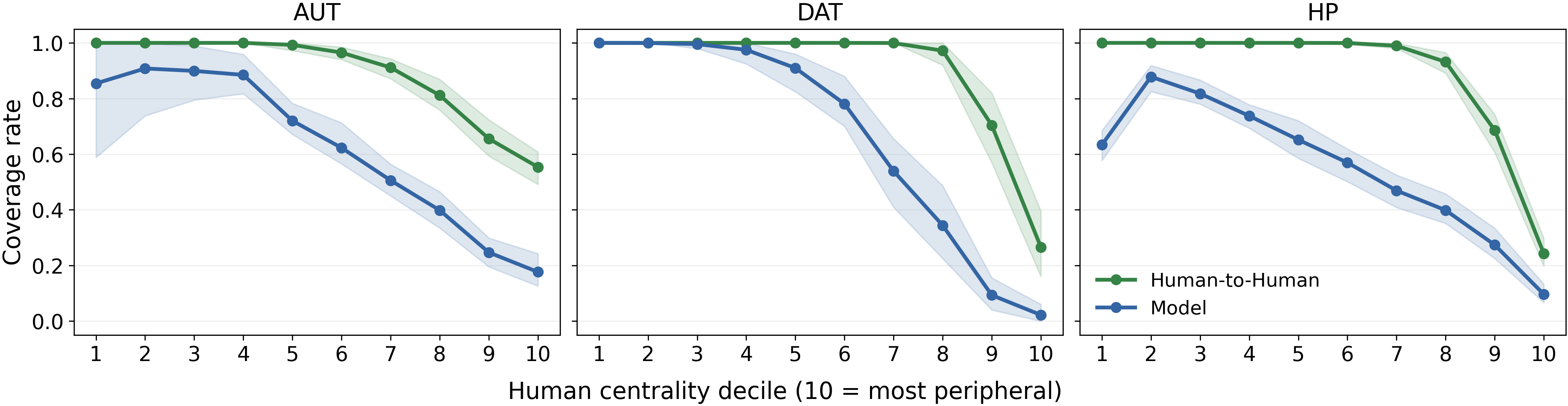}
\caption{\textbf{Model coverage is disproportionately central relative to the Human-to-Human reference.}}
\label{fig:centrality-coverage}
\end{figure*}

Figure~\ref{fig:centrality-coverage} gives a distributional interpretation of "average" generation: model reach disproportionately favors central human responses. Coverage declines toward the periphery for both models and held-out humans, but models fall further behind the human benchmark in these regions. Across all three tasks, the model-to-human coverage ratio is lower in the outer deciles than in the task overall.

\subsection{What Models Reach and Miss}
\label{sec}

We next examine how covered and uncovered HP fanfiction passages differ in content and form. Using one split Qwen2.5-7B at $t{=}1.0$, the highest-coverage HP configuration, we identify 1,085 covered and 915 uncovered passages in one experiment. We compare lexical and stylistic features alongside LLM-judge annotations of character and relationship content. Feature definitions and annotation details appear in \ref{app:operationalization}.

\begin{table}[t]
\centering
\footnotesize
\setlength{\tabcolsep}{0.5pt}
\renewcommand{\arraystretch}{1.13}
\begin{tabular}{l l c r}
\toprule
\textbf{Axis} & \textbf{Variable} & \textbf{Diff.} & $\boldsymbol{p}$ \\
\midrule
\multirow{4}{*}{\makecell[l]{Canonical\\anchoring}}
& Character present & $+0.296$ & $<0.001$ \\
& Character mentions / 1k & $+10.99$ & $<0.001$ \\
& Setting present & $+0.197$ & $<0.001$ \\
& Canon vocab present & $+0.115$ & $<0.001$ \\
\midrule
\multirow{3}{*}{\makecell[l]{Style and\\rhythm}}
& Quote span count & $+2.50$ & $<0.001$ \\
& Dialogue ratio & $+0.065$ & $<0.001$ \\
& Fragment ratio & $-0.073$ & $<0.001$ \\
\midrule
\multirow{3}{*}{\makecell[l]{Fandom\\context}}
& Fanon vocab present & $-0.015$ & $0.220$ \\
& Non-HP characters & $-0.095$ & $<0.001$ \\
& Implicit HP world & $+0.073$ & $<0.001$ \\
\midrule
\multirow{3}{*}{\makecell[l]{Relationship\\content}}
& Romance present & $+0.009$ & $0.679$ \\
& M/M relationship & $-0.030$ & $0.170$ \\
& M/F relationship & $+0.063$ & $0.002$ \\
\bottomrule
\end{tabular}
\vspace{0.1cm}
\caption{\textbf{Descriptive differences between covered and uncovered HP passages.}
Diff.\ is Covered minus Uncovered: proportion differences for binary features
and mean differences for continuous features.}
\label{tab:cov-uncov-summary}
\end{table}

We conduct statistical comparisons of textual features between the covered and uncovered passage sets (Table~\ref{tab:cov-uncov-summary}). Covered texts more often explicitly mention elements established in the original Harry Potter book series, including recognizable characters (e.g., Harry, Hermione, and Draco), locations (e.g., Hogwarts and Diagon Alley), and series-specific terms (e.g., Quidditch, Auror, and Horcrux). They also contain more quoted dialogue. Uncovered texts contain fewer such references, more often introduce characters from outside the Harry Potter series.

Beyond these surface-level lexical and stylistic markers, we also examine character relationships. M/F (Male and Female) relationships are more common in covered passages, with the same direction across model-specific comparisons. M/M (Male and Male) relationships are slightly less common in the covered set.

This case study shows how researchers can identify patterns in the human texts models miss, even when the writing is varied and difficult to categorize. These patterns can suggest directions for improvement, while our coverage metrics can help assess whether changes to models, prompts, or sampling strategies reach previously missed responses. We outline potential extensions in the Discussion as well.

\subsection{Pairwise Audit of Geometric Matches}
We audit 600 nearest-neighbor pairs across AUT, DAT, and HP using judgments from two human annotators and two LLM judges. Table~\ref{tab:annotation-audit} reports rates of \emph{clearly similar} judgments averaged across the four evaluations for human--model and human--human pairs. Appendix~\ref{app:pair-audit} details the protocol, judge identities, and agreement analysis (Fleiss' $\kappa=.490$).

\begin{table}[t]
\centering
\scriptsize
\setlength{\tabcolsep}{4pt}
\begin{tabular}{@{}llccc@{}}
\toprule
& & \multicolumn{2}{c}{Human--model} & Human--human \\
\cmidrule(lr){3-4}\cmidrule(l){5-5}
Task & Similarity & Uncovered & Covered & Nearest \\
\midrule
AUT & Use/function/action & .23 & .77 & .83 \\
DAT & Concept relation & .14 & .46 & .39 \\
HP & Character/setting & .19 & .45 & .50 \\
HP & Theme & .32 & .52 & .62 \\
HP & Event & .11 & .32 & .42 \\
HP & Relationship & .22 & .40 & .57 \\
HP & Style & .24 & .43 & .70 \\
\bottomrule
\end{tabular}
\caption{\textbf{Strict-match rates averaged across two human annotations and two disclosed LLM-judge evaluations.}}
\label{tab:annotation-audit}
\end{table}
% this is annotated by me and gemini 3.8 flash and codex 6.1, dont know if we could claim like this.

Across AUT, DAT, and HP, the audit shows a consistent pattern: covered pairs are more often judged clearly similar, providing strong support for our geometric coverage metric. HP also shows that proximity alone may not capture all narrative details, making covered–uncovered analysis useful for examining these differences more closely. Together, these results demonstrate the framework’s value in measuring coverage and investigating what models miss. We discuss possible extensions in the discussion part.

\section{Discussion}

Across AUT, DAT, and Harry Potter fanfiction, the tested LLMs generally produce in-boundary responses but cover substantially less of the human reference distribution than equally sized human samples. Under the primary global geometry, their reach is disproportionately central. We interpret this as a distributional sense of "average" generation: model samples more readily reach shared human patterns while leaving more peripheral variation uncovered. Our approach to analyzing LLM coverage allows us to measure this tendency and identify the types of human responses that are covered or missed. % makes this tendency measurable and identifies the human responses behind it.

Our use of the HP fanfiction task enables textual interpretation of the gaps we uncover. Covered human passages contain more explicit canon names and places and more dialogue, while uncovered passages more often contain non-HP characters and sentence fragments. The pairwise audit further shows that a nearby model passage can share a theme while differing in events or style. Together, these analyses concretely connect the observed coverages gap to differences in the content and form of the writing.

The framework allows researchers to compare models, prompts, or decoding strategies at a fixed sample budget and inspect whether improvements reach previously uncovered responses. A possible extension is free-text survey analysis. When responses combine multiple ideas or values, sparse autoencoders could extract thematic features from their embeddings. Applying geometric coverage in this shared feature space could help researchers examine which themes and combinations of ideas model samples reach or miss. Once these features are learned and validated, reusing them across comparisons could reduce the cost of repeated LLM annotation.

Finally, these results provide empirical content to our notion of “cultural reach.” The term is relative: it describes not how much of a culture a model captures, but how much of a community’s observed range of expression its outputs reach compared with an equally sized sample of that community’s own writers. By this measure, the models we test reach roughly half as much of Harry Potter fanfiction as human writers do, and their reach is weighted toward what the fandom already shares. That AUT and DAT show the same shape without a community suggests a general property of model sampling, one that takes on cultural significance where a community’s variety is at stake. This has broader stakes because, with increasing widespread use of LLMs to create culture, such tendencies have potentially problematic downstream effects by making some forms of expression to encounter and reproduce easier than others, whether stories, song lyrics, or social media posts. That said, broader coverage of course is not automatically desirable. It is up to researchers and the communities themselves to decide. Our method simply helps them make such decisions in a more informed and empirical way.

\section{Limitations}

Our framework has several limitations. First, it relies on sentence-level embeddings to construct the empirical human response boundary. These embeddings capture broad semantic similarity, but may miss dimensions important for open-ended cultural generation, such as narrative structure, voice, genre convention, pragmatic nuance, and community-specific meaning. Furthermore, different embedding methods or hybrid representations could yield different boundary estimates.

Second, the representativeness of our boundary is constrained by the quality of our human data. Our datasets provide concrete reference distributions, but they are finite and cannot capture the full range of human variation. Larger and more diverse human samples may reveal additional regions of human response space that LLMs cover or miss.

Third, our boundary depends on a neighborhood-radius choice. We use \(p50\) for consistency and report robustness checks, but absolute IBR and LLM-Cov values can vary with this threshold. Future work could explore adaptive or task-specific boundary selection.

Fourth, our HP fanfiction analysis is diagnostic rather than exhaustive. The covered-versus-uncovered features are proxies for richer forms of fandom-specific meaning, and we do not claim that the same patterns generalize to all open-ended domains. For example, although the mixed-gender relationship direction is consistent across model-specific splits, its magnitude and statistical reliability vary across configurations.

\begingroup
\sloppy
\newpage

\bibliography{references_revised}
\endgroup

\appendix

\newpage
% =================================================================
% APPENDIX
% =================================================================
\clearpage
\section{Relative Coverage Across Configurations}
\label{app:relative-coverage-summary}

LLM-Cov and matched H2H-Cov are the primary reported coverage quantities. Table~\ref{tab:relative-coverage-summary} summarizes $R$ across the ten model--temperature configurations for each task.

\begin{table}[h]
\centering
\small
\begin{tabular}{lcc}
\toprule
Task & Mean $R$ & Range of $R$ \\
\midrule
AUT & .396 & [.106, .699] \\
DAT & .432 & [.100, .745] \\
HP & .527 & [.364, .624] \\
\bottomrule
\end{tabular}
\caption{Relative coverage across configurations under the global $p50$ radius. Configuration values summarize the 50 repeated splits. The range describes differences among configurations, not a population confidence interval.}
\label{tab:relative-coverage-summary}
\end{table}

\section{Reproducibility Details}
\label{app:reproducibility}

This section documents the prompts, models, decoding settings, and
sampling conventions used throughout the paper. Our goal is to make
every coverage measurement reported in the main text reproducible
from publicly available models without ambiguity about generation
parameters, prompt formatting, or human reference construction.
Section~3 summarizes the end-to-end pipeline,
from raw human and model responses to the in-boundary rate (IBR) and
LLM-coverage (LLM-Cov) metrics.

% \begin{figure}[h]
% \centering
% \includegraphics[width=0.95\linewidth]{image/flowchart.png}
% \caption{\textbf{End-to-end coverage pipeline.} Human and model
% responses are embedded with the same sentence encoder and projected
% into a shared PCA space. The empirical human response boundary is
% the union of $\varepsilon$-balls around human embeddings, where
% $\varepsilon$ is set as a quantile of human $k$-nearest-neighbor
% distances. IBR is the fraction of model outputs that fall inside this
% boundary; LLM-Cov is the fraction of human responses reached by at
% least one model output.}
% \label{fig:flowchart-app}
% \end{figure}

\subsection{Task Prompts}
\label{app:prompts}

For each task we use a fixed prompt template applied identically
across all models and temperatures, holding prompt wording constant
across these comparisons. Prompts were
designed to elicit the same kind of response from LLMs that humans
give in the corresponding psychometric task, while keeping the output
format easy to parse.

\paragraph{Alternative Uses Task (AUT).}
AUT is a divergent-thinking task in which participants list creative
uses for an everyday object. We instantiate it with the object
\texttt{rope} and ask the model to produce a numbered list of short
verb phrases.

\begin{quote}
\small\itshape
You are helping brainstorm creative uses. Generate \{per\_call\}
distinct, feasible uses for the object `\{object\_name\}'. Respond
with 1--5 word verb phrases only (no full sentences, no extra nouns),
number the list 1 through N, and avoid repeats or explanations.
\end{quote}

Each item is then normalized into the canonical frame ``A potential
use of the rope is \ldots'' before sentence embedding, which prevents
trivial surface differences in how the model formats list items from
affecting embedding geometry.

\paragraph{Divergent Association Task (DAT).}
DAT asks participants to produce a set of mutually unrelated common
nouns; semantic distance among the chosen words is taken as an index
of divergent thinking. We use the standard 10-word version:

\begin{quote}
\small\itshape
Please enter 10 words that are as different from each other as
possible in meaning and usage. Rules: (1) Only single words in
English. (2) Only common nouns. (3) No proper nouns. (4) No
specialized or technical vocabulary. (5) Think of the words on your
own, do not reference anything you can currently see. Return the
words as a numbered list 1--10, one noun per line, with no
explanations.
\end{quote}

\paragraph{HP fanfiction.}
For the Harry Potter experiment we use a length-controlled short prompt
implemented. It specifies only the
fandom, target length, and story-only output format; it does not request
familiar characters, settings, plot structure, pairings, or tropes.

\begin{quote}
\small\itshape
\textbf{System.} You are a creative fiction writer.\\

\textbf{User.} Write a Harry Potter fanfiction story in 200 to 250
words. Output only the story.

\end{quote}

Outputs that fell outside the 200--250-word window were discarded.
Human excerpts are also filtered to 200--250 words. The two sets therefore
share the same permitted length range, although their length distributions
need not be identical.

\subsection{HP Corpus--Prompt Comparability}
\label{app:hp-comparability}

The HP comparison does not use identically elicited human and model texts.
Model outputs are complete short stories produced from the prompt above and
accepted only at 200--250 words. Human texts are cleaned AO3 story-body
excerpts selected at 200--250 words; they were written for a community
publishing context rather than in response to the model prompt. Consequently,
differences in fragmentation, exposition, and narrative closure can reflect
both model behavior and the corpus--prompt mismatch. We interpret the HP
results conditionally: they describe which portions of this AO3 reference
sample are reached by outputs from this generation protocol. The current data
do not separate model effects from elicitation and excerpt-selection effects.

\subsection{Models, Decoding, and Generation Counts}
\label{app:models}

We evaluate five open-source instruction-tuned LLMs spanning three
model families (Qwen, Llama, and Mistral/Ministral) and a 7B--32B
parameter range (Table~\ref{tab:models-app}). All models are publicly
available through HuggingFace. The local generation helper uses
family-specific prompt formatting: the Llama header format, Mistral
\texttt{[INST]} format, and a plain System/User/Assistant format for
the remaining models.

\begin{table*}[t]
\caption{\textbf{Model identifiers used for generation.}}
\label{tab:models-app}
\centering
\small
\setlength{\tabcolsep}{4pt}
\renewcommand{\arraystretch}{1.15}
\begin{tabular}{ll}
\toprule
\textbf{Short name} & \textbf{HuggingFace identifier} \\
\midrule
Qwen2.5-7B   & \texttt{Qwen/Qwen2.5-7B-Instruct} \\
Qwen2.5-32B  & \texttt{Qwen/Qwen2.5-32B-Instruct} \\
Llama-3.1-8B & \texttt{meta-llama/Llama-3.1-8B-Instruct} \\
Ministral-8B & \texttt{mistralai/Ministral-8B-Instruct-2410} \\
Mistral-24B  & \texttt{MistralAI/Mistral-Small-24B-Instruct-2501} \\
\bottomrule
\end{tabular}
\end{table*}

All main coverage results use two decoding temperatures, $t = 0.3$ and
$t = 1.0$. In the HP generation, decoding uses multinomial
sampling with $\texttt{top\_p}=0.9$, repetition penalty 1.1, and at
most 360 newly generated tokens. Generation is performed one story per
call. A story is accepted only when its whitespace-tokenized length is
between 200 and 250 words; generation continues until 2,000 accepted
stories are obtained for the configuration. The final manifest contains
2,000 valid and unique stories for each of the ten HP
model--temperature configurations, or 20,000 HP model outputs in total.

The two temperatures and five models give ten model--temperature
configurations per task. We keep configuration sizes fixed within each
task when comparing models and temperatures. Within each model-to-human
coverage calculation, the model generator and human target contain the
same number of responses.

\subsection{Human Reference Construction}
\label{app:human-ref}

\begin{table*}[t]
\centering
\small
\setlength{\tabcolsep}{5pt}
\renewcommand{\arraystretch}{1.08}
\begin{tabular}{p{0.08\textwidth}p{0.27\textwidth}p{0.22\textwidth}p{0.29\textwidth}}
\toprule
\textbf{Task} & \textbf{Public source} & \textbf{Source files} & \textbf{Selection used in this study} \\
\midrule
AUT &
Multilingual AUT Repository;
\url{https://osf.io/5cy9n/} &
\texttt{english2.csv}--\texttt{english6.csv} &
Responses to the \texttt{rope} prompt were retained and normalized,
yielding 4,479 eligible responses; 4,000 were used in the matched analysis. \\

DAT &
Divergent Association Task repository;
\url{https://osf.io/vjazn/} &
\texttt{study2.tsv} from the Data component
(\url{https://osf.io/kbeq6/}) &
Each record contains one participant's complete ten-word response.
We use 1,000 response records in the matched analysis. \\

HP &
Archive of Our Own; &
Original and supplementary work-URL lists retained &
Downloaded story bodies were cleaned, deduplicated, and filtered to
200--250 words, yielding 4,435 eligible excerpts; 4,000 were used in
the matched analysis. \\
\bottomrule
\end{tabular}
\caption{\textbf{Human-data sources and selection.}
Counts in the final column distinguish the eligible source pool from
the analysis sample used for matched reference and held-out assignments.}
\label{tab:human-data-sources}
\end{table*}

The empirical human response boundary is constructed from a human
pool whose size and provenance differ by task. For AUT we
use 4,000 human responses to the \texttt{rope} prompt, and for DAT we
use 1,000 response records. Each AUT record is one normalized rope use.
Each DAT record is one complete normalized participant response containing
the requested word list; the words are embedded jointly rather than treated
as separate evaluation items. For HP fanfiction, we successfully download
4,486 AO3 text files from the original and supplementary URL lists.
Each file is cleaned by retaining story-body text between AO3 body
markers and excluding metadata, notes, chapter headings, and files
matching a predefined list of obvious non-story or crossover
indicators. To accommodate small differences between AO3's displayed
count and our whitespace tokenizer, we accept cleaned bodies containing
200--250 whitespace-separated words. This yields 4,435 unique eligible
excerpts. We do not balance the sample by author.

For each task, the Human-to-Human reference uses two non-overlapping,
equally sized human samples, one serving as the reference set and the
other as the held-out generator set. Thus, within every H2H calculation,
the human generator and target are also equal in size. The
reference set fits the representation and defines the boundary.
H2H-IBR is the fraction of query humans lying within that boundary,
whereas H2H-Cov is the fraction of reference humans reached by at least
one query human. This produces two complete H2H measurements without
self-coverage.

\subsection{Sample Ledger and Evaluation Units}
\label{app:sample-ledger}

Table~\ref{tab:sample-ledger} records the counts used by every main matched
comparison. ``Human total'' is the pool assigned without overlap to the
reference and held-out roles. The model query is truncated to the held-out
count. The 50 assignments reuse the same finite pool with seeds
42000--42049, so they measure assignment sensitivity rather than 50
independent populations.

\begin{table*}[t]
\caption{\textbf{Sample ledger for the matched main protocol.} Counts are
numbers of evaluation units, not pairwise distances.}
\label{tab:sample-ledger}
\centering
\small
\setlength{\tabcolsep}{5pt}
\begin{tabular}{lp{0.35\textwidth}rrrr}
\toprule
Task & Evaluation unit & Human total & Reference & Held-out & Model query \\
\midrule
AUT & One normalized response giving a use for rope & 4,000 & 2,000 & 2,000 & 2,000 \\
DAT & One normalized complete word-list response & 1,000 & 500 & 500 & 500 \\
HP & One cleaned AO3 excerpt or one generated short story & 4,000 & 2,000 & 2,000 & 2,000 \\
\bottomrule
\end{tabular}
\end{table*}

The available AUT file retains response text but no participant identifier;
the DAT file retains a response identifier but no producer identifier. These
tasks therefore use response-level assignments. The HP analysis file retains
work identifiers but not author identifiers. Work-level exact duplicates are
removed, but author-level partitioning cannot be reconstructed from the
current files.

\subsection{HP Coverage Computation}
\label{app:hp-coverage-computation}

We encode the HP reference humans, held-out humans, and model passages with
\path{sentence-transformers/all-mpnet-base-v2}, using normalized
embeddings and a maximum encoder sequence length of 384 tokens. PCA is
fit only on the 2,000 reference-human embeddings. We retain the smallest
number of components explaining at least 90\% of reference-human
variance, capped at 200. Held-out-human and model embeddings are centered
and projected using this reference-human-fitted transformation. For the fixed split used in the HP diagnostic analysis, the resulting
space has 159 dimensions and explains 0.9010 of the reference-human
variance.

For each reference-human embedding, we compute the Euclidean distance
to its 15th nearest distinct reference human. The global radius
$\varepsilon$ is the median ($p50$) of these distances; in the fixed
HP split, $\varepsilon=0.782970$. IBR is the
fraction of a configuration's 2,000 model outputs whose nearest
reference-human distance is no greater than $\varepsilon$. Consistent
with the AUT and DAT protocol, LLM-Cov is the fraction of these same
2,000 reference humans whose nearest model-output distance is no greater
than $\varepsilon$. The non-overlapping held-out human set is used only
for the Human-to-Human reference: H2H-Cov measures the fraction of the
reference humans reached by the held-out humans, and H2H-IBR measures
the reverse direction. The HP covered-vs-uncovered diagnostic uses only
the highest-coverage single configuration, Qwen2.5-7B at $t=1.0$; it
does not pool temperatures or take a multi-model union.

\section{Robustness and Budget Analyses}
\label{app:robustness}

We rerun the robustness analyses with the same data and matched-split
protocol as the main results.  For each of 50 assignments, the reference
humans alone fit PCA and define the neighborhood radius; the model sample and
held-out-human sample have equal size.  All summaries below average the
per-split values over the ten model--temperature configurations unless a
specific configuration is named.  The 50 assignments overlap and therefore
measure sensitivity to the finite reference construction, not population
uncertainty.

\subsection{Encoder Robustness}
\label{app:encoder-current}

We compare MPNet with MiniLM and BGE-M3 while retaining PCA90, $k=15$, and
the global $p50$ radius.  Each encoder uses L2-normalized embeddings with a
maximum sequence length of 384 tokens and fits its own reference-only PCA and
radius within each assignment.  Table~\ref{tab:encoder-current} reports mean H2H
coverage, mean model coverage, the mean ratio $R$, and the highest
configuration-level mean $R$.  Every one of the 90 encoder--task--configuration
combinations has mean $R<1$.

\begin{table*}[t]
\caption{\textbf{Matched coverage under three text encoders.} Means are over
50 assignments and ten model--temperature configurations. ``Best $R$'' is the
largest configuration-level mean within a task and encoder.}
\label{tab:encoder-current}
\centering
\small
\setlength{\tabcolsep}{5pt}
\begin{tabular}{llcccc}
\toprule
Encoder & Task & H2H-Cov & LLM-Cov & Mean $R$ & Best $R$ \\
\midrule
MPNet   & AUT & .889 & .352 & .396 & .699 \\
        & DAT & .894 & .386 & .432 & .745 \\
        & HP  & .885 & .467 & .527 & .624 \\
MiniLM  & AUT & .881 & .347 & .394 & .665 \\
        & DAT & .949 & .396 & .417 & .727 \\
        & HP  & .899 & .530 & .590 & .703 \\
BGE-M3  & AUT & .844 & .283 & .336 & .583 \\
        & DAT & .956 & .477 & .499 & .792 \\
        & HP  & .952 & .557 & .584 & .694 \\
\bottomrule
\end{tabular}
\end{table*}

Absolute coverage changes with the representation, especially for DAT and
HP, but the calibrated comparison is stable: model samples cover less of the
reference set than equally sized human samples under every encoder.  We do not
interpret this as showing that one encoder is intrinsically superior; it shows
that the aggregate finding is not specific to MPNet.

\subsection{PCA, Boundary-Quantile, and $k$ Sensitivity}
\label{app:geometry-current}

Table~\ref{tab:geometry-current} varies one MPNet geometry choice at a time.
Because a looser radius raises both model and human coverage, the relevant
quantity is their calibrated ratio rather than either raw coverage alone.

\begin{table*}[t]
\caption{\textbf{Mean relative human coverage under geometry choices.} Each
cell is mean $R$ over 50 assignments and ten configurations.  The main setting
is PCA90, $p50$, and $k=15$.}
\label{tab:geometry-current}
\centering
\small
\setlength{\tabcolsep}{5pt}
\begin{tabular}{llccc}
\toprule
Choice & Setting & AUT & DAT & HP \\
\midrule
PCA variance & 85\% & .395 & .438 & .517 \\
             & 90\% & .396 & .432 & .527 \\
             & 95\% & .391 & .409 & .532 \\
\midrule
Radius quantile & $p25$ & .205 & .273 & .355 \\
                & $p50$ & .396 & .432 & .527 \\
                & $p75$ & .595 & .627 & .700 \\
                & $p90$ & .758 & .819 & .822 \\
\midrule
$k$ & 5  & .286 & .299 & .381 \\
    & 10 & .356 & .377 & .467 \\
    & 15 & .396 & .432 & .527 \\
    & 20 & .431 & .470 & .569 \\
    & 30 & .506 & .523 & .623 \\
\bottomrule
\end{tabular}
\end{table*}

The radius quantile and $k$ affect the magnitude, as expected, but mean $R$
remains below one throughout.  At the most permissive tested radius, $p90$,
even the best configuration in each task remains below its H2H reference
(best $R=.928$ for AUT, $.928$ for DAT, and $.856$ for HP).  The PCA result is
comparatively insensitive over 85--95\% retained variance.

\subsection{Matched Sampling-Budget Curves}
\label{app:budget-current}

We vary the model and held-out-human generator budgets
together while keeping the fitted reference geometry fixed within a split.
For each configuration, samples are nested: a larger budget extends the
smaller sample rather than drawing an unrelated set.  Table~\ref{tab:budget-current}
reports the best main-paper configuration for each task at 10\%, 50\%, and
100\% of the available matched generator budget.  The final column computes
the ratio within the three most peripheral deciles.

\begin{table*}[t]
\caption{\textbf{Coverage as matched model and human budgets increase.}
Values are means over 50 assignments for Ministral-8B at $t=1.0$ on AUT and
Qwen2.5-7B at $t=1.0$ on DAT and HP.}
\label{tab:budget-current}
\centering
\small
\setlength{\tabcolsep}{4pt}
\begin{tabular}{lrrrrcc}
\toprule
Task & Fraction & Budget & H2H-Cov & LLM-Cov & $R$ & $R_{\rm outer}$ \\
\midrule
AUT & .1 & 200  & .631 & .327 & .518 & .507 \\
    & .5 & 1000 & .831 & .529 & .637 & .405 \\
    & 1.0 & 2000 & .889 & .622 & .699 & .407 \\
DAT & .1 & 50   & .699 & .527 & .755 & .325 \\
    & .5 & 250  & .853 & .632 & .742 & .236 \\
    & 1.0 & 500 & .894 & .666 & .745 & .237 \\
HP  & .1 & 200  & .657 & .351 & .535 & .613 \\
    & .5 & 1000 & .837 & .496 & .593 & .427 \\
    & 1.0 & 2000 & .885 & .553 & .624 & .413 \\
\bottomrule
\end{tabular}
\end{table*}

More samples increase raw model and H2H coverage, but do not close the
matched-budget gap over the observed range.  The peripheral ratio is also
lower than the overall ratio at the full budget for every task.  These are
finite-range observations; the table does not establish an asymptotic ceiling
or imply that coverage could never converge with substantially more samples.

\subsection{Combining Model Configurations at Fixed and Growing Budgets}
\label{app:ensemble-current}

We also pool all ten model--temperature configurations in two ways.  Under the
fixed-total policy, the task's original model budget is divided as evenly as
possible across the ten configurations.  Under the growing-union policy, all
outputs from every configuration are retained, producing ten times as many
model outputs as the matched human generator.  Table~\ref{tab:ensemble-current}
therefore separates gains from configuration breadth from gains due to a
larger total sample.

\begin{table*}[t]
\caption{\textbf{All-configuration coverage under fixed and growing model
budgets.} H2H uses the task's original matched generator budget.}
\label{tab:ensemble-current}
\centering
\small
\setlength{\tabcolsep}{4pt}
\begin{tabular}{lrrrrrrrr}
\toprule
Task & H2H & Mean single & Best single & Fixed $n_M$ & Fixed mix & Fixed $R$ & Growing $n_M$ & Growing $R$ \\
\midrule
AUT & .889 & .352 & .622 & 2000 & .561 & .631 & 20000 & .837 \\
DAT & .894 & .386 & .666 & 500  & .676 & .756 & 5000  & .887 \\
HP  & .885 & .467 & .553 & 2000 & .544 & .615 & 20000 & .785 \\
\bottomrule
\end{tabular}
\end{table*}

At a fixed total budget, the ten-configuration mixture exceeds the mean
single-configuration coverage on every task. It exceeds the best single
configuration only on DAT, by .010; it is lower by .061 on AUT and .009 on
HP. Configuration breadth therefore does not yield a consistent advantage
over selecting the best observed configuration. Retaining all outputs raises
coverage further, but that comparison gives the model ten times the human
generator budget and is descriptive rather than a matched-budget test or a
single-model continuation-sampling control.

\subsection{Internal Diversity and Human Coverage}
\label{app:diversity-current}

For each configuration we compute mean pairwise cosine distance in the
original normalized encoder space.  Table~\ref{tab:diversity-current} gives
the Spearman correlation between that internal-diversity statistic and mean
$R$ across the ten configurations.  The two quantities are related, but their
rankings are not identical, especially for MPNet on HP.  Internal diversity
also lacks an H2H target and cannot localize which reference responses are
missed.

\begin{table*}[t]
\caption{\textbf{Descriptive correlation between internal diversity and
relative human coverage.} Each coefficient uses ten model--temperature
configurations; no population inference is attached to these small-$n$
correlations.}
\label{tab:diversity-current}
\centering
\small
\begin{tabular}{lccc}
\toprule
Encoder & AUT & DAT & HP \\
\midrule
MPNet  & .685 & .661 & .345 \\
MiniLM & .673 & .891 & .648 \\
BGE-M3 & .794 & .745 & .782 \\
\bottomrule
\end{tabular}
\end{table*}

\section{HP Fanfiction Covered-vs-Uncovered Diagnostics}
\label{app:cov-uncov-extended}

The main paper presents a focused covered-vs-uncovered comparison on
HP fanfiction. This appendix reports the lexical, stylistic, and
relationship features measured for the same highest-coverage single
configuration and documents how each feature is operationalized.

\subsection{Setup}
\label{app:hp-setup}

We define the primary \textsc{Covered} and \textsc{Uncovered} groups
using the highest-coverage short-prompt configuration, Qwen2.5-7B at
$t{=}1.0$. A human excerpt is \textsc{Covered} if at least one of this
configuration's 2,000 outputs lies within $\varepsilon$ of it. The
target is the same 2,000-reference-human set used to fit PCA and define
the boundary, matching the AUT and DAT protocol. Of these excerpts,
1,085 (54.3\%) are \textsc{Covered} and 915 (45.8\%) are
\textsc{Uncovered}. No temperatures or models are pooled in this
diagnostic analysis.

\subsection{Full Variable Comparison}
\label{app:full-variable-comparison}

Table~\ref{tab:cov-uncov-full} reports the principal features we measure,
organized into canonical anchoring, style and rhythm, lexical
variation, and relationship orientation. Continuous variables report group
means with Mann--Whitney $U$ $p$-values; binary indicators report
group rates with chi-square $p$-values. The $\Delta$ column gives
the covered-minus-uncovered difference, so positive values mean a
feature is more common in covered excerpts. Table~\ref{tab:hp-bootstrap-intervals}
reports percentile bootstrap intervals for the contrasts selected for the
main-text summary; the machine-readable output retains the corresponding
resamples and statistics.

\begin{table*}[t]
\caption{\textbf{Full covered-vs-uncovered comparison} for HP
fanfiction. $\Delta$ is Covered minus Uncovered: positive values
indicate features more common in covered excerpts, and negative
values indicate features more common in uncovered excerpts.
The largest differences concern canonical anchoring and dialogue.}
\label{tab:cov-uncov-full}
\centering
\small
\setlength{\tabcolsep}{4pt}
\renewcommand{\arraystretch}{1.13}
\begin{tabular}{p{0.42\textwidth} c c c r}
\toprule
\textbf{Variable} & \textbf{Cov.} & \textbf{Uncov.} & $\boldsymbol{\Delta}$ & $\boldsymbol{p}$ \\
\midrule
\multicolumn{5}{l}{\textbf{Canonical anchoring}} \\[1pt]

HP setting present (rate)        & 0.372 & 0.175 & $+$0.197 & $2.3{\times}10^{-22}$ \\
HP setting / 1k words (mean)     & 2.87  & 1.23  & $+$1.64  & $5.0{\times}10^{-24}$ \\
Canon vocab present (rate)       & 0.336 & 0.222 & $+$0.115 & $2.0{\times}10^{-8}$ \\
Canon vocab / 1k words (mean)    & 2.60  & 1.53  & $+$1.07  & $4.7{\times}10^{-10}$ \\
HP character present (rate)      & 0.891 & 0.596 & $+$0.296 & $1.0{\times}10^{-52}$ \\
HP character / 1k words (mean)   & 31.11 & 20.12 & $+$10.99 & $4.5{\times}10^{-38}$ \\

\midrule
\multicolumn{5}{l}{\textbf{Style and rhythm}} \\[1pt]

Dialogue ratio (mean)            & 0.317 & 0.253 & $+$0.065  & $2.0{\times}10^{-12}$ \\
Quoted spans (mean)              & 8.10 & 5.59 & $+$2.50   & $1.9{\times}10^{-24}$ \\
Fragment ratio (mean)            & 0.264 & 0.337 & $-$0.073  & $7.3{\times}10^{-5}$ \\

\midrule
\multicolumn{5}{l}{\textbf{Lexical variation}} \\[1pt]

Type--token ratio (mean)         & 0.613 & 0.566 & $+$0.047 & $1.3{\times}10^{-10}$ \\
Fanon vocab present (rate)       & 0.064 & 0.079 & $-$0.015 & $0.220$ \\

\midrule
\multicolumn{5}{l}{\textbf{Fandom-internal signals}} \\[1pt]

Non-HP characters (rate)         & 0.098 & 0.192 & $-$0.095 & $2.0{\times}10^{-9}$ \\
Implicit HP world (rate)         & 0.999 & 0.926 & $+$0.073 & $9.9{\times}10^{-19}$ \\

\midrule
\multicolumn{5}{l}{\textbf{Relationship orientation}} \\[1pt]

Romance present (rate)           & 0.726 & 0.717 & $+$0.009 & $0.679$ \\
Family-centered (rate)           & 0.319 & 0.261 & $+$0.058 & $0.006$ \\
Friendship-centered (rate)       & 0.288 & 0.249 & $+$0.039 & $0.055$ \\
M/M relationship (rate)          & 0.327 & 0.357 & $-$0.030 & $0.170$ \\
M/F relationship (rate)          & 0.320 & 0.257 & $+$0.063 & $0.002$ \\
F/F relationship (rate)          & 0.077 & 0.091 & $-$0.013 & $0.323$ \\
Ship tension present (rate)      & 0.478 & 0.473 & $+$0.005 & $0.855$ \\

\bottomrule
\end{tabular}
\end{table*}

\begin{table*}[t]
\caption{\textbf{Uncertainty intervals for the HP contrasts summarized in
the main text.} Differences are Covered minus Uncovered. Brackets give
percentile 95\% bootstrap intervals for this fixed diagnostic sample; they do
not account for author dependence or repeated feature selection.}
\label{tab:hp-bootstrap-intervals}
\centering
\scriptsize
\setlength{\tabcolsep}{4pt}
\begin{tabular}{lrr@{\qquad}lrr}
\toprule
Feature & Difference & 95\% interval & Feature & Difference & 95\% interval \\
\midrule
Character present & $+.296$ & $[.259,.331]$ & Quote spans & $+2.50$ & $[1.91,3.08]$ \\
Character mentions / 1k & $+10.99$ & $[8.70,13.24]$ & Dialogue ratio & $+.065$ & $[.045,.084]$ \\
Setting present & $+.197$ & $[.160,.235]$ & Fragment ratio & $-.073$ & $[-.093,-.053]$ \\
Canon vocabulary present & $+.115$ & $[.075,.154]$ & Fanon vocabulary present & $-.015$ & $[-.038,.008]$ \\
Non-HP characters & $-.095$ & $[-.126,-.064]$ & Implicit HP world & $+.073$ & $[.057,.091]$ \\
Romance present & $+.009$ & $[-.030,.049]$ & M/M relationship & $-.030$ & $[-.072,.012]$ \\
M/F relationship & $+.063$ & $[.023,.103]$ & & & \\
\bottomrule
\end{tabular}
\end{table*}

The full table reinforces the canonical-anchoring result. Covered
excerpts are more likely to name HP characters and settings and use
canon-specific vocabulary. Dialogue and quoted spans also strongly
separate the groups. The type--token ratio is higher and the fragment
ratio lower in covered excerpts, whereas fanon-vocabulary presence does
not differ significantly. Family-centered passages and M/F relationships
are more common in covered excerpts. Romance, friendship, M/M, F/F, and
ship-tension rates do not differ at the raw two-sided $p<.05$ threshold.
These results describe the best short-prompt configuration only.

\subsection{Model-Specific Relationship Check}
\label{app:model-specific-relationship}

To assess whether the mixed-gender relationship direction is unique to
the primary configuration, we repeat the covered--uncovered comparison for each
short-prompt model at $t=1.0$. Table~\ref{tab:mf-by-model} shows that
the covered-minus-uncovered difference is positive for all five models,
although its magnitude and statistical reliability vary.

\begin{table*}[t]
\caption{\textbf{Model-specific mixed-gender relationship check.}
Each row uses one short-prompt model at $t=1.0$ to define covered and
uncovered reference excerpts.}
\label{tab:mf-by-model}
\centering
\small
\setlength{\tabcolsep}{5pt}
\renewcommand{\arraystretch}{1.13}
\begin{tabular}{lcccc}
\toprule
\textbf{Model} & \textbf{Cov. rate} & \textbf{Uncov. rate} & $\boldsymbol{\Delta}$ & $\boldsymbol{p}$ \\
\midrule
Qwen2.5-7B   & 0.320 & 0.257 & $+$0.063 & $0.002$ \\
Qwen2.5-32B  & 0.317 & 0.261 & $+$0.056 & $0.007$ \\
Llama-3.1-8B & 0.313 & 0.266 & $+$0.048 & $0.022$ \\
Ministral-8B & 0.320 & 0.262 & $+$0.057 & $0.005$ \\
Mistral-24B  & 0.301 & 0.281 & $+$0.021 & $0.331$ \\
\bottomrule
\end{tabular}
\end{table*}

\subsection{Variable Operationalization}
\label{app:operationalization}

Table~\ref{tab:operationalization} documents how each variable group
is operationalized. The canonical-anchoring and style features are
deterministic, computed by rule-based matchers and text statistics
applied uniformly across both groups. We report Mann--Whitney $U$ tests
for continuous features and two-sided chi-square tests for binary
features, comparing the 1,085 best-configuration-covered and 915
best-configuration-uncovered reference excerpts. Relationship labels
are produced for the same reference-human split by a
deterministic (greedy-decoding) Mistral-Small-24B judge.

\begin{table*}[t]
\caption{\textbf{Variable operationalization} for the HP
covered-vs-uncovered diagnostics.}
\label{tab:operationalization}
\centering
\small
\setlength{\tabcolsep}{4pt}
\renewcommand{\arraystretch}{1.20}
\begin{tabular}{p{0.18\textwidth} p{0.38\textwidth} p{0.37\textwidth}}
\toprule
\textbf{Group} & \textbf{Variables} & \textbf{Operationalization} \\
\midrule
Canonical anchoring &
\texttt{setting\_present}\newline
\texttt{setting\_count}\newline
\texttt{setting\_per\_1k} &
Rule-based match against canonical HP setting terms, such as
Hogwarts, Gryffindor, Slytherin, Diagon Alley, Hogsmeade, Ministry,
Grimmauld Place, Privet Drive, Burrow, and Azkaban. \\
Canon vocabulary &
\texttt{canon\_vocab\_present}\newline
\texttt{canon\_vocab\_count}\newline
\texttt{canon\_vocab\_per\_1k} &
Rule-based match against HP-world terms, such as wand, spell,
potion, Quidditch, Auror, Death Eater, Horcrux, Patronus, Animagus,
Muggle, pureblood, and Order. \\
Characters &
\texttt{character\_present} &
Rule-based match against an enumerated list of HP character names
and surnames. \\
Style and rhythm &
\texttt{word\_count}\newline
\texttt{sentence\_count}\newline
\texttt{mean\_sentence\_len}\newline
\texttt{sentence\_len\_var}\newline
\texttt{fragment\_ratio}\newline
\texttt{dialogue\_ratio}\newline
\texttt{quote\_span\_count}\newline
\texttt{punctuation\_density}\newline
\texttt{ellipsis\_density} &
Deterministic text statistics from sentence segmentation, word
tokenization, quoted spans, punctuation counts, ellipses, and
sentence-length distributions. \\
Affect and narrative function &
\texttt{comfort\_*}\newline
\texttt{angst\_*}\newline
\texttt{subversion\_*} &
Rule-based keyword proxies for affective and narrative-function cues. \\
Fandom-internal signals &
\texttt{characters\_outside\_hp}\newline
\texttt{hp\_world\_implicit} &
Binary LLM-judge annotations for whether an excerpt introduces non-HP
characters and whether it presupposes the HP world without explicit
exposition. The same judge prompt and deterministic decoding are
applied to both coverage groups. \\
Relationship orientation &
\texttt{relationship\_type\_romance}\newline
\texttt{relationship\_type\_family}\newline
\texttt{relationship\_type\_friendship}\newline
\texttt{relationship\_gender\_mm}\newline
\texttt{relationship\_gender\_mf}\newline
\texttt{relationship\_gender\_ff}\newline
\texttt{ship\_tension\_present} &
Binary LLM-judge labels for romantic, family, and friendship content;
the gender composition of the primary romantic relationship; and explicit
ship tension. The annotator receives the passage without its coverage label,
and the same prompt and deterministic decoding are applied to both groups. \\
\bottomrule
\end{tabular}
\end{table*}

The relationship prompt defines \texttt{relationship\_type\_family} as
content centered on kinship, parent--child or sibling relations, ancestry,
household relations, or family reconciliation and conflict. It defines
\texttt{relationship\_gender\_mf} only when the primary romantic or ship-coded
relation is between a male and a female character. These labels may overlap
with other relationship labels; they are binary passage attributes rather than
mutually exclusive genres.

The deterministic features are independent of any annotation model and
can be reproduced exactly given the keyword lists. The group comparisons are descriptive of this fixed
reference split and model configuration; they do not identify causal effects
or population-level category biases.

\section{Matched-Split and Radius Robustness}
\label{app:matched-robustness}

The matched analysis uses the following evaluation totals: 4,000 AUT responses, 1,000 DAT responses, and a 4,000-response HP analysis sample. In each repetition, the total is randomly assigned without overlap to equal reference and held-out roles; model samples match the held-out count. We repeat this procedure with seeds 42000--42049. PCA and every radius are fitted from the reference assignment only. The same fitted representation and threshold are then used for the model and held-out human comparisons. These repeated, overlapping assignments quantify split sensitivity rather than population uncertainty.

\begin{table*}[t]
\caption{\textbf{Global-radius matched coverage across 50 assignments.} The best row values identify the highest-$R$ configuration for each task. Mean and range summarize all ten model--temperature configurations.}
\label{tab:global-matched-app}
\centering
\small
\begin{tabular}{llccccc}
\toprule
Task & Best configuration & H2H & LLM & Best $R$ & Mean $R$ & Range \\
\midrule
AUT & Ministral-8B, $t{=}1.0$ & .889 & .622 & .699 & .396 & [.106,.699] \\
DAT & Qwen2.5-7B, $t{=}1.0$ & .894 & .666 & .745 & .432 & [.100,.745] \\
HP & Qwen2.5-7B, $t{=}1.0$ & .885 & .553 & .624 & .527 & [.364,.624] \\
\bottomrule
\end{tabular}
\end{table*}

We also replace the shared radius $\varepsilon$ with a point-specific radius $r_i$, defined for each reference response by its 15th distinct nearest reference neighbor. Coverage then requires a query sample to fall within $r_i$ of reference response $i$. This follows the local support construction used by prior precision/recall and coverage estimators more closely.

\begin{table*}[t]
\caption{\textbf{Point-specific local-radius robustness across 50 assignments.} DAT and HP H2H coverage is at or near one, showing a ceiling effect.}
\label{tab:local-matched-app}
\centering
\small
\begin{tabular}{lcccc}
\toprule
Task & Best H2H & Best LLM & Best $R$ & Mean $R$ \\
\midrule
AUT & .889 & .578 & .650 & .351 \\
DAT & 1.000 & .809 & .809 & .397 \\
HP & ${\sim}1.000$ & .593 & .593 & .493 \\
\bottomrule
\end{tabular}
\end{table*}

All ten configurations in all three tasks have mean $R<1$ under both radius definitions; under the global definition, every task--configuration pair also has $R<1$ in every one of the 50 assignments. The local estimator supports the aggregate model--human comparison but compresses the human baseline on DAT and HP. This is why the shared global radius remains the primary operationalization.

\subsection{Centrality Definition and Repeated-Assignment Bands}
\label{app:centrality-definition}

Centrality is computed separately within each reference assignment. Let
$\bar d_{15}(h_i)$ be the mean distance from reference response $h_i$ to its
15 nearest distinct reference neighbors, $d_{15}(h_i)$ the distance to the
15th neighbor, $d_c(h_i)=\lVert h_i-\bar h\rVert_2$ its distance from the
reference centroid, and $\operatorname{LOF}_{15}(h_i)$ its Local Outlier
Factor score. We define the peripheral score
\[
\begin{aligned}
s_i={}&z(\bar d_{15}(h_i))+z(d_{15}(h_i))+z(d_c(h_i))\\
     &+z(\operatorname{LOF}_{15}(h_i))
      -z\!\left(\frac{1}{\bar d_{15}(h_i)+10^{-8}}\right),
\end{aligned}
\]
where every $z(\cdot)$ is standardized within the current reference set.
Stable ranks of $s_i$ are divided into ten equal-sized bins; decile 1 is most
central and decile 10 most peripheral. This is an embedding-space composite,
not a measure of social frequency or cultural importance.

For Figure~\ref{fig:centrality-coverage}, each line is the mean decile
coverage over 50 overlapping assignments. The shaded band is the 2.5th--97.5th
percentile of those assignment-level rates. It summarizes sensitivity to the
reference assignment and is not a population confidence interval. Duplicate
AUT vectors can create ties in local scores; the stable ranking rule makes the
partition deterministic given an assignment.

\begin{table*}[t]
\caption{\textbf{Coverage in the three most peripheral embedding-space deciles under the global $p50$ radius.} All values are means over 50 assignments for the task's best configuration.}
\label{tab:outer-deciles-app}
\centering
\small
\begin{tabular}{lcccc}
\toprule
Task & H2H-Cov & LLM-Cov & $R_{\mathrm{outer}}$ & Splits with model lower \\
\midrule
AUT & .674 & .274 & .407 & 50/50 \\
DAT & .647 & .153 & .237 & 50/50 \\
HP & .620 & .256 & .413 & 50/50 \\
\bottomrule
\end{tabular}
\end{table*}

\section{Pairwise Audit of Geometric Matches}
\label{app:pair-audit}

\paragraph{Purpose and fixed configuration.}
The audit examines which task-relevant properties are captured by embedding distance and the coverage threshold. It does not ask the judge to redefine coverage and does not validate IBR. We fix seed 42000 before sampling and use Ministral-8B at $t=1.0$ for AUT and Qwen2.5-7B at $t=1.0$ for DAT and HP. All pairs use MPNet--PCA, $k=15$, and the global $p50$ threshold fitted from that split's reference humans.

\paragraph{Sampling.}
For each task, we sample 100 distinct reference responses: 25 each from covered-near, covered-far, uncovered-near, and uncovered-far strata, where near/far is defined within coverage status by normalized nearest-neighbor distance. Each reference is paired once with its nearest model response and once with its nearest held-out human response. This yields 200 pairs per task and 600 total. The same candidate may be nearest to multiple references, and this reuse is retained in the mapping. Equal stratum counts are a diagnostic sample and are not interpreted as population prevalence without sampling weights.

\paragraph{Blind annotation.}
Pair order and A/B order are randomized, and source type, model identity,
embedding distance, coverage status, and sampling stratum are hidden from all
evaluators. AUT pairs are judged according to whether the two responses
describe the same use, function, or action; mentioning the shared object
\emph{rope} alone is not sufficient. DAT pairs are judged according to
conceptual overlap or relations between the embedded response units. HP pairs
are judged separately along five dimensions: character or setting, theme,
specific event, relationship interaction, and expression style. These
dimension-specific criteria distinguish thematic or narrative overlap from
similarity in concrete events or manner of expression.

Each dimension receives one of four labels: \emph{not similar},
\emph{partly similar}, \emph{clearly similar}, or \emph{cannot judge}.
Evaluators may additionally provide short textual evidence for their
judgments. Two human annotators and two LLM judges independently evaluate the
same 600 frozen pairs. The LLM judges are \texttt{gpt-6.1-sol} and \texttt{gemini 3.8 flash}. All four evaluators receive the same task-specific rubric and blinded
pair presentation. The annotation packets hide source type, model identity,
embedding distance, coverage status, sampling stratum, and the other
evaluators' labels.

\paragraph{Human annotation participation.}
Graduate student volunteers were directly invited by the authors to participate
in the annotation. Participation was voluntary, and no monetary compensation was provided. The annotators provided written consent for their judgments to be used in research. We do not publicly disclose identifying information about them.

\paragraph{Instructions for human annotators.}
Human annotators were instructed to judge only the two displayed responses
and not to infer whether either response was written by a human or generated
by a model. Differences in wording alone were not treated as evidence that
two responses were dissimilar. For each applicable dimension, annotators
selected one of four labels: \emph{not similar}, \emph{partly similar},
\emph{clearly similar}, or \emph{cannot judge}. A pair was
\emph{clearly similar} when it shared a recognizable and substantive match
on the specified dimension, and \emph{partly similar} when the match was
meaningful but incomplete.

For AUT, annotators judged whether the two responses expressed the same use,
function, or action. Because every response concerned a rope, merely referring
to the same object was insufficient for similarity. For DAT, annotators judged
conceptual overlap or a meaningful relation between the two complete
ten-word sets. A single incidentally shared word would normally support, at
most, a judgment of partial similarity.

For HP fanfiction, annotators evaluated five dimensions separately.
\emph{Character or setting} concerned overlap in recognizable characters,
character roles, locations, or story-world situations. \emph{Theme} concerned
the substantive idea or concern developed by the passages; generic references
to magic, love, friendship, or conflict were insufficient for clear
similarity. \emph{Specific event} concerned whether the passages shared a
recognizable event or event structure, without requiring identical characters,
wording, or outcomes. \emph{Relationship interaction} concerned the dynamic
between characters, such as protection, rivalry, reconciliation, mentoring,
or family conflict. \emph{Expression style} concerned how the passages were
written, including dialogue, viewpoint, pacing, fragmentation, tone, and
narrative presentation.

We do not publicly disclose identifying information about the human annotator. Source texts, identifying metadata, text-containing annotation packets, and private pair mappings will not be publicly redistributed.

\paragraph{Analysis.}
The strict summary treats only \emph{clearly similar} as a match;
\emph{partly similar} and \emph{not similar} are treated as nonmatches, and
\emph{cannot judge} is excluded for the corresponding dimension.
Table~\ref{tab:pair-audit} averages the four strict-match judgments within the
human--model subset. The Spearman column correlates normalized embedding
distance with the pair-level mean ordinal judgment, coding \emph{not similar},
\emph{partly similar}, and \emph{clearly similar} as 0, 1, and 2,
respectively. Across all task judgments, overall Fleiss'
$\kappa$ is .490. Detailed agreement outputs are released with the analysis.

\paragraph{Human-only direction check.}
We repeat the covered--uncovered comparison using only the two human label
sets. For both human annotators, the strict-match rate is higher among covered
than uncovered human--model pairs in all seven task--dimension comparisons
(Table~\ref{tab:human-only-pair-audit}). Across the 1,400 binary
task--dimension judgments, the two human annotators have 74.4\% raw agreement
and Cohen's $\kappa=.492$. This check shows that the covered--uncovered
direction in the combined audit does not depend on the two LLM judges.

\begin{table*}[!t]
\caption{\textbf{Strict-match rates within human--model pairs.}
Only ``clearly similar'' counts as a match. Rates are averaged across two
human annotations and two disclosed LLM-judge evaluations. Difference is
covered minus uncovered. Negative $\rho$ indicates that greater normalized
embedding distance is associated with a weaker mean ordinal judgment.}
\label{tab:pair-audit}
\centering
\small
\setlength{\tabcolsep}{7pt}
\begin{tabular}{llrrrr}
\toprule
Task & Audited dimension & Uncovered & Covered & Difference & Spearman $\rho$ \\
\midrule
AUT & Same use/function/action & .23 & .77 & +.54 & $-.701$ \\
DAT & Conceptual relation & .14 & .46 & +.32 & $-.652$ \\
HP & Character/setting & .19 & .45 & +.26 & $-.434$ \\
HP & Theme & .32 & .52 & +.20 & $-.492$ \\
HP & Specific event & .11 & .32 & +.22 & $-.588$ \\
HP & Relationship interaction & .22 & .40 & +.19 & $-.484$ \\
HP & Expression style & .24 & .43 & +.19 & $-.526$ \\
\bottomrule
\end{tabular}

\par\bigskip

\caption{\textbf{Human-only strict-match check within human--model pairs.}
Only ``clearly similar'' counts as a match. Human A and Human B columns give
each annotator's covered-minus-uncovered difference. Mean rates average the
two human annotations.}
\label{tab:human-only-pair-audit}
\centering
\small
\setlength{\tabcolsep}{6pt}
\begin{tabular}{llrrrrr}
\toprule
Task & Audited dimension & Human A $\Delta$ & Human B $\Delta$ & Mean uncov. & Mean cov. & Mean $\Delta$ \\
\midrule
AUT & Same use/function/action & +.46 & +.56 & .25 & .76 & +.51 \\
DAT & Conceptual relation & +.26 & +.22 & .01 & .25 & +.24 \\
HP & Character/setting & +.12 & +.34 & .12 & .35 & +.23 \\
HP & Theme & +.26 & +.04 & .24 & .39 & +.15 \\
HP & Specific event & +.22 & +.10 & .06 & .22 & +.16 \\
HP & Relationship interaction & +.22 & +.08 & .27 & .42 & +.15 \\
HP & Expression style & +.24 & +.14 & .37 & .56 & +.19 \\
\bottomrule
\end{tabular}
\end{table*}

\paragraph{Interpretive boundary.}
The audit supports the claim that covered human--model pairs more often contain
clearly recognizable task-relevant similarity and that larger embedding
distance generally corresponds to weaker judged similarity. It does not show
complete semantic equivalence. The moderate overall agreement shows that
boundary cases remain judgment-sensitive; the audit therefore supports an
approximate interpretation of geometric reach rather than treating coverage
as a human-defined equivalence relation.

\end{document}